\documentclass[conference]{IEEEtran}

\usepackage[paper=letterpaper,top=54pt,bottom=54pt,left=54pt,right=54pt]{geometry}
\usepackage{afterpage}
\usepackage{cite}
\usepackage{amsmath,amssymb,amsfonts}
\usepackage{algorithmic}
\usepackage{graphicx}
\usepackage{textcomp}
\usepackage{xcolor}
\usepackage{graphicx}
\usepackage{comment}
\usepackage{subcaption}
\usepackage{siunitx}
\usepackage{float}
\usepackage{color}
\usepackage{multirow}
\usepackage{xcolor} 
\usepackage{amssymb}
\usepackage{pifont}
\usepackage{ragged2e}
\newcommand{\cmark}{\ding{51}}
\newcommand{\xmark}{\ding{55}}
\usepackage{tablefootnote}
\include{bib_short.def}
\usepackage{adjustbox}
\usepackage{tabularx}
\definecolor{yellow}{RGB}{145,77,56}

\def\BibTeX{{\rm B\kern-.05em{\sc i\kern-.025em b}\kern-.08em
    T\kern-.1667em\lower.7ex\hbox{E}\kern-.125emX}}

\begin{document}

\noindent
\begin{minipage}{\textwidth}
\centering
\huge 
\textbf{IEEE Copyright Notice} \\[0.8em] 
\normalsize
© 2025 IEEE. Personal use of this material is permitted. Permission from IEEE must be obtained for all \\
other uses, in any current or future media, including reprinting/republishing this material for advertising \\
or promotional purposes, creating new collective works, for resale or redistribution to servers or lists, or \\
reuse of any copyrighted component of this work in other works.\\
This work has been accepted for publication at IEEE IV2025. The final published version will be available via IEEE Xplore.
\end{minipage}

\vspace{1.5cm}

\title{BEV-LLM: Leveraging Multimodal BEV Maps for Scene Captioning in Autonomous Driving
}
\newgeometry{top=72pt, bottom=54pt, left=54pt, right=54pt}
\afterpage{\aftergroup\restoregeometry}
\author{\IEEEauthorblockN{Felix Brandstätter\IEEEauthorrefmark{1}, Erik Schütz\IEEEauthorrefmark{1}, Katharina Winter\IEEEauthorrefmark{1}, and Fabian B. Flohr\IEEEauthorrefmark{1}}
\IEEEauthorblockA{\IEEEauthorrefmark{1}Intelligent Vehicles Lab (IVL)\\
Munich University of Applied Sciences, 80335 Munich, Germany\\
Email: intelligent-vehicles@hm.edu}
}

\maketitle

\begin{abstract}
Autonomous driving technology has the potential to transform transportation, but its wide adoption depends on the development of interpretable and transparent decision-making systems. Scene captioning, which generates natural language descriptions of the driving environment, plays a crucial role in enhancing transparency, safety, and human-AI interaction. We introduce BEV-LLM, a lightweight model for 3D captioning of autonomous driving scenes. BEV-LLM leverages BEVFusion to combine 3D LiDAR point clouds and multi-view images, incorporating a novel absolute positional encoding for view-specific scene descriptions. 
Despite using a small 1B parameter base model, BEV-LLM achieves competitive performance on the nuCaption dataset, surpassing state-of-the-art by up to 5\% in BLEU scores.
Additionally, we release two new datasets — nuView (focused on environmental conditions and viewpoints) and GroundView (focused on object grounding) — to better assess scene captioning across diverse driving scenarios and address gaps in current benchmarks, along with initial benchmarking results demonstrating their effectiveness.
\end{abstract}


\section{INTRODUCTION}
\label{sec:introduction}

Autonomous driving technology has the potential to revolutionize transportation systems, but its widespread adoption hinges on the development of interpretable and explainable decision-making. Scene captioning, the generation of natural language descriptions of the environment, has emerged as a research effort towards higher transparency and safety of autonomous vehicles.
Facilitating more natural human-AI interaction, scene captioning allows users to query the system about its surroundings and decision-making \cite{sima2023drivelm,xu2024drivegpt4, malla2023drama, hwang2024emma}.

Recent advances in computer vision and natural language processing have paved the way for a more comprehensive scene understanding in autonomous driving contexts. The integration of multiple sensor modalities, particularly the fusion of camera images and LiDAR data, has shown promising results in creating a holistic environment representation. Bird's-Eye View (BEV)~\cite{liu2023bevfusion} representations have gained popularity due to their ability to provide a unified perspective of the scene, incorporating spatial relationships and object interactions.

LiDAR-LLM~\cite{2023_Yang_LiDAR_LLLM_Exploring_the_Potential_of_Large_Language_Models_for_3D_LiDAR_Understanding} pushed the state-of-the-art in 3D captioning of outdoor scenes, but shows limitations using LiDAR as a single modality and the lack of absolute positional information.

\begin{figure}[t]
    \centering
    \def\svgwidth{\linewidth}
    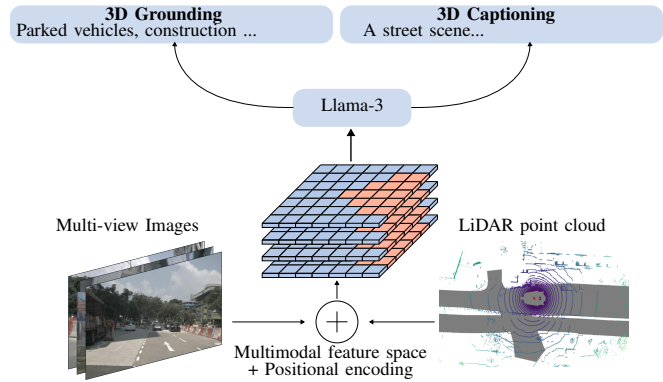 
    \caption{We present BEV-LLM, our 3D captioning model for autonomous driving scenes. Our proposed Positional Q-Former takes BEV maps from fused LiDAR pointclouds and surround-view images as inputs and appends an absolute positional encoding, which leverages the reasoning capabilities of the LLM to generate view-specific scene descriptions and grounding.}
    \label{fig:BEV-LLM_motivation}
\end{figure}

Our proposed method, BEV-LLM, builds upon the recent success of LiDAR-LLM by adding camera images as another modality input combined with a new positional encoding using absolute positional vectors. We employ a pre-trained BEVFusion model \cite{liu2023bevfusion} to fuse the six surround-view images and the LiDAR pointcloud from nuScenes \cite{2020_Caesar_nuScenes_A_Multimodal_Dataset_for_Autonomous_Driving}, preserving semantically rich and geometric features crucial for robust information retrieval in perception tasks for autonomous driving.
Figure~\ref{fig:BEV-LLM_motivation} shows our approach of applying our positional encoding to the BEV feature maps to generate view-specific 3D captioning and grounding descriptions with LLama-3.

\afterpage{\aftergroup\restoregeometry}
Our contributions are threefold.
First, we leverage BEVFusion~\cite{liu2023bevfusion} to enhance scene captioning by combining 360-degree multi-view images and LiDAR data into a unified BEV representation. This fusion enriches captions by capturing both environmental context from images and precise spatial details from LiDAR, outperforming single-modality methods like LiDAR-LLM~\cite{2023_Yang_LiDAR_LLLM_Exploring_the_Potential_of_Large_Language_Models_for_3D_LiDAR_Understanding}.
Second, we propose a sine-cosine positional embedding that partitions the feature space into six views based on input images. Conditioned on user input, it enables flexible focus on specific views or the full scene, surpassing simpler masking techniques~\cite{2023_Yang_LiDAR_LLLM_Exploring_the_Potential_of_Large_Language_Models_for_3D_LiDAR_Understanding}. Third, BEV-LLM achieves competitive performance in the 3D captioning task using a lightweight 1B parameter model, outperforming competitors on the nuCaption dataset with a BLEU-2 score of 30.02\%, a BLEU-3 of 24.42\% and BLEU-4 of 20.28\%.
To better assess scene captioning across driving scenarios and conditions, we introduce
two new datasets: \textit{nuView} (focused on environmental conditions and viewpoints) and \textit{GroundView} (focused on object grounding), along with initial benchmarking results.
Datasets are available on our website\footnote{https://iv.ee.hm.edu/bev-llm}; models and source code will be released.

\section{RELATED WORK}
\label{sec:relatedwork}
\subsection{Large Language Models (LLMs) for AD}

LLMs have become increasingly prevalent in autonomous driving due to their ability to process natural language, enabling enhanced scene understanding and generalization to unseen scenarios. These models are used in various aspects of autonomous driving, from perception and planning to human-machine interaction and explainability.

Many recent end-to-end architectures integrate LLMs to reason about scenes and generate future trajectories~\cite{sima2023drivelm, jiang2024senna}. 
For instance, DriveLM~\cite{sima2023drivelm} uses Graph VQA to reason about scenes by answering questions related to different stages of the autonomous driving pipeline. 
Senna~\cite{jiang2024senna} predicts trajectories and describes scenes based on multi-view images.

The integration of natural language also facilitates human interaction, enhancing explainability and interpretability. DriveGPT4~\cite{xu2024drivegpt4} pioneered end-to-end motion planning with LLMs, providing descriptions and justifications of the ego-vehicle's actions. This research has been extended by works integrating explanations into trajectory planning~\cite{yuan2024rag, wang2023drivemlm} and utilizing scene descriptions for enhanced explainability~\cite{hwang2024emma}.

\subsection{Scene Captioning and Visual Question Answering}

Scene Captioning and Visual Question Answering (VQA) have emerged as crucial research areas in autonomous driving, aiming to provide detailed descriptions or answer specific queries about the driving scene to enhance LLM's scene understanding and decision-making.

Earlier approaches in this field focused on 2D image-based methods. These include models for caption generation in risk communication~\cite{mori2021image, malla2023drama}, VQA frameworks for explaining driving actions~\cite{atakishiyev2023explaining}, and CLIP-based traffic scene captioning~\cite{zhang2024tsic}. ADAPT~\cite{jin2024adapt} unified autonomous driving and video captioning with real-time decision reasoning.

Recent advancements have shifted towards BEV representations, offering a more comprehensive perspective of the driving scene. BEV-TSR~\cite{bevtsr2024} introduced a framework that leverages descriptive text inputs to retrieve corresponding scenes in the BEV space based on camera images, addressing challenges in global feature representation and text retrieval in complex driving scenarios.

LiDAR-LLM~\cite{2023_Yang_LiDAR_LLLM_Exploring_the_Potential_of_Large_Language_Models_for_3D_LiDAR_Understanding} pioneered the use of LiDAR data for 3D scene descriptions, introducing the nuCaption and nuGrounding datasets. Although this approach advanced 3D captioning for autonomous driving, it faces limitations due to its reliance on LiDAR data alone.

Our proposed approach, BEV-LLM, addresses these limitations by integrating camera images, as the BEV map is capable of easily unifying different sensor setups into a consistent representation, and employing a more efficient positional encoding. 
By using Llama-3.2 for improved language modeling and incorporating multi-view positional embeddings, BEV-LLM generates richer and more accurate scene descriptions, enhancing LLM-based training with positional context.

Concurrently with our research, $\text{TOD}^3$Cap~\cite{jin2025tod3cap} was introduced as a powerful 3D captioning method for outdoor scenes, also addressing the modality gap in LiDAR-LLM. A detailed comparison with $\text{TOD}^3$Cap remains a topic for future work.

\section{METHOD}
\label{sec:method}
\subsection{Architecture Overview}

\begin{figure*}[h]
    \centering
    \def\svgwidth{0.72\linewidth}
    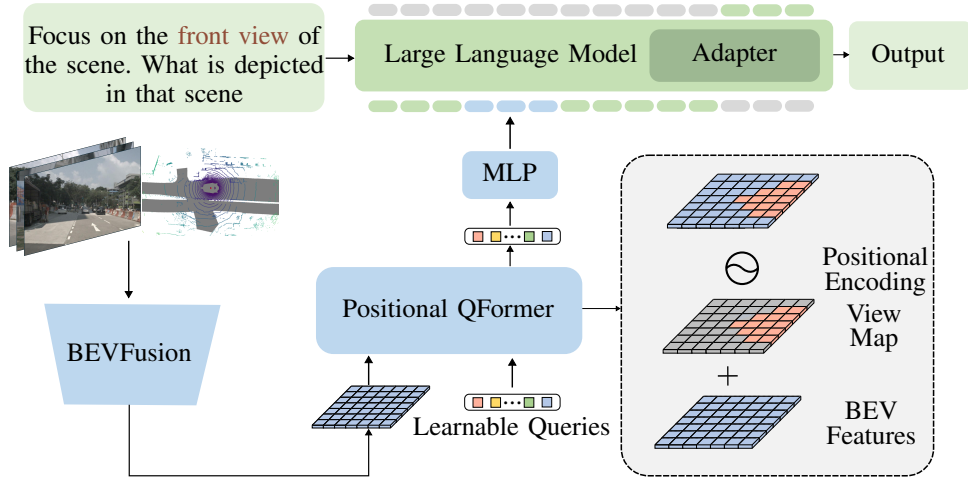
    \caption{Overall architecture of BEV-LLM. LiDAR pointclouds and camera images are encoded using BEVFusion. The Positional Q-Former appends the BEV features with positional information and aligns them with the text feature space before an MLP projects them into the embedding space of the LLM, together with the tokenized text embeddings from the prompt. A LoRA adapter is integrated to fine-tune the pre-trained LLM on the 3D captioning task, while the LLM is kept frozen.}
    \label{fig:BEV-LLM_architecture}
\end{figure*}

The architecture of BEV-LLM is depicted in Figure \ref{fig:BEV-LLM_architecture}.
We employ a pre-trained BEVFusion model~\cite{liu2023bevfusion} to fuse the six surround-view images with a resolution of 1600 x 900 and the LiDAR pointcloud from nuScenes \cite{2020_Caesar_nuScenes_A_Multimodal_Dataset_for_Autonomous_Driving} for each sample into a flattened BEV feature map of size $\left[512, 32400\right]$. Combining dense image data with sparse but spatially precise LiDAR features preserves semantically rich and geometric features.

The BEV features are aligned with a convolution-based BEV encoder and flattened to shape $\left[512, 32, 400\right]$.

We propose the Positional Q-Former as a novel adaptation of the Q-Former \cite{2023_Li_BLIP_2_Bootstrapping_Language_Image_Pre_Training_with_Frozen_Image_Encoders_and_Large_Language_Models}  processing BEV maps instead of visual images and encoding them positionally, which is described in detail in Subsection (\ref{subsec:q-former}).
The features generated by the frozen BEVFusion encoder are transformed by the Positional Q-Former using $512$ learnable queries with dimension $768$. 
For the 3D captioning task, the extracted features from the Positional Q-Former are projected into the embedding space of the LLM with an MLP with output shape $\left[512, 2048\right]$ and concatenated with the tokenized text embeddings to form the input to the LLM.swer. 

We integrate Meta's LLama3-8B-Instruct and LLama3.2-1B-Instruct as LLM, which are both models from the LLama3 family with the least amount of parameters and pre-trained on large-scale text data with further training on instruction tuning data 
\cite{dubey2024llama}.

\subsection{Positional Q-Former} \label{subsec:q-former}
We adapt the Q-Former as introduced by Li et al. \cite{2023_Li_BLIP_2_Bootstrapping_Language_Image_Pre_Training_with_Frozen_Image_Encoders_and_Large_Language_Models} to bridge the modality gap between BEV features and text embeddings. 
Instead of the image transformer from the original Q-Former architecture, the Positional Q-Former contains BEVFusion, and the learnable queries interact with the frozen BEV features via cross-attention. With a dimension of $768$, the queries reduce the incoming BEV feature maps that have dimensions $\left[512\times32,400\right]$.

We increase the queries in the Positional Q-Former to $512$ due to the higher number of input patches from the BEV encoder ($32,400$) compared to the ViT-L/14 ($257$) that is used for the Q-Former \cite{2023_Li_BLIP_2_Bootstrapping_Language_Image_Pre_Training_with_Frozen_Image_Encoders_and_Large_Language_Models}.
Our Positional Q-Former implementation is initialized with pre-trained weights from $\text{BERT}_\text{base}$ \cite{2019_Devlin_BERT_Pre_Training_of_Deep_Bidirectional_Transformers_for_Language_Understanding}.

\begin{figure}[ht]
    \centering
    \def\svgwidth{\linewidth}
    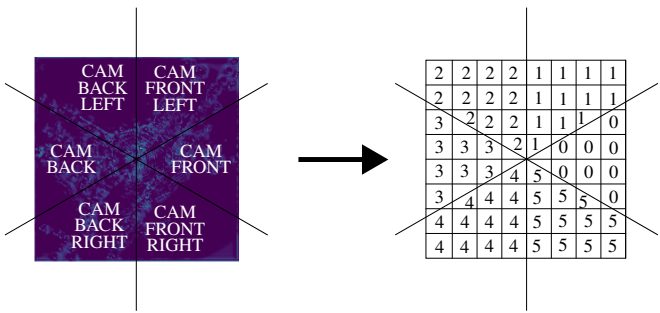 
    \caption{The BEV map is split into six equal sectors corresponding to the camera views available in the nuScenes dataset by assigning every cell in the BEV map to a view position $v\in\left[0,1,2,3,4,5\right]$ to enable view-aware processing.}
    \label{fig:bev-map-splitting}
\end{figure}

In autonomous driving, view-awareness is key to accurately grounding objects in the surrounding scene and reasoning about them. 
To retain positional awareness corresponding to the different camera views in the aligned feature space, we introduce a novel positional encoding to the BEV feature maps. 
While Yang et al. \cite{2023_Yang_LiDAR_LLLM_Exploring_the_Potential_of_Large_Language_Models_for_3D_LiDAR_Understanding} provide $c$-dimensional, learnable positional embeddings $V_p\in\mathrm{R}^{c\times6}$ for LiDAR-LLM, we treat the views as fixed geometric relationships, which is more parameter efficient and maintains fixed geometrical relationships for precise spatial reasoning, avoiding spurious correlations that could occur with learned embeddings.
For the positional encoding, we create a $180\times180$ sized view-classification map congruent with the computed BEV feature maps. As depicted in Figure \ref{fig:bev-map-splitting}, the map is divided into six views centered around the ego vehicle's position with every cell assigned to a view position $v\in\left[0,1,2,3,4,5\right]$, each corresponding to one of the cameras in the nuScenes dataset \cite{2020_Caesar_nuScenes_A_Multimodal_Dataset_for_Autonomous_Driving}.

We encode the view position of each cell with the sinusoidal embedding function used by Vaswani et al. \cite{2017_Vaswani_Attention_is_all_you_need} to a 512-dimensional vector, resulting in a feature map of size $\left[512\times180\times180\right]$. 

\subsection{Prompting Scheme}
We apply the prompting scheme from LLama-3, which provides characteristic tokens for the start and end of a message, as well as special header tokens defining the role attributed to the subsequent message, which is either \textit{user}, \textit{system} or \textit{assistant}. Question and answer tokens train the LLM for the VQA task, with the question providing an instruction and the answer given by the ground truth scene description.
We design a conversational prompt between the user and the assistant, in which the user role is the querying instance providing a BEV-LLM specific $\mathrm{<}$bev$\mathrm{>}$ token for the insertion of the Positional Q-Former queries. 
In the response, the \textit{assistant} header token indicates the beginning of the response to be generated by the LLM. The ground truth response containing the dataset caption is provided during training and omitted during inference.

\section{EXPERIMENTS}
\subsection{Settings}
\label{sec:settings}

We propose two versions of BEV-LLM with different LLM and Positional Q-Former configurations:
BEV-LLM-8B incorporates Llama-3-8B-Instruct, an 8 billion parameter instruction-tuned language model. This setup utilizes a Positional Q-Former with 256 queries.
BEV-LLM-1B  integrates Llama-3.2-1B-Instruct, the 1 billion parameter variant, coupled with a Positional Q-Former featuring 512 queries.

\subsection{Training}
\label{sec:training}
The pre-trained BEVFusion module and Llama language model remain frozen during training. The Positional Q-Former, the MLP and the LoRA \cite{hu2021lora} adapter are jointly trained on 8 NVIDIA A100 GPUs for six epochs with a total training time of 48 hours and a batch size of 3 for BEV-LLM-1B and 72 hours for BEV-LLM-8B but only on 4 GPUs with a batch size of 2 due to storage restrictions.
With this fine-tuning strategy, we efficiently train the model end-to-end while preserving the foundational knowledge embedded in the pre-trained models. To train the Positional Q-Former for the tasks of scene capturing, we use a threefold loss: BEV-Text Contrastive Learning (BTC) of positive and negative query and text pairs, BEV-grounded Text Generation (BTG) for BEV feature queries to attend text representations, and BEV-Text Matching (BTM) to align BEV maps and text with binary classification.

\subsection{Datasets}
\label{sec:dataset}
We train and evaluate BEV-LLM on the following three datasets with different focuses on each: environmental, traffic, and object-level focus, leading to different types of scene descriptions.
The size of the datasets, intended tasks, and the focus of the scene descriptions are summarized in Table \ref{tab:datasets}. 

\begin{table}[h!]
\centering
\LARGE
\resizebox{\linewidth}{!}{
\begin{tabular}{|c|c|c|c|c|c}
\hline
\textbf{Dataset} & \textbf{\# Samples (t/v)} & \textbf{3D Captioning} & \textbf{Obj. Grounding} & \textbf{Focus} \\
\hline
nuCaption & 240k (169k/70k) & \cmark & \xmark & Traffic \\
\hline
nuView& 205k (169k/36k) & \cmark & \xmark & Environment \\
\hline
GroundView & 7.4k (6k/1.4k) & \xmark & \cmark & Objects \\
\hline
\end{tabular}}
\caption{Datasets overview listing sample size including train and validation splits, the intended tasks: 3D Captioning and Object-Level Grounding, and the main focus of the dataset.}
\label{tab:datasets}
\end{table}

\begin{figure}[h]
    \centering
    \def\svgwidth{\linewidth}
    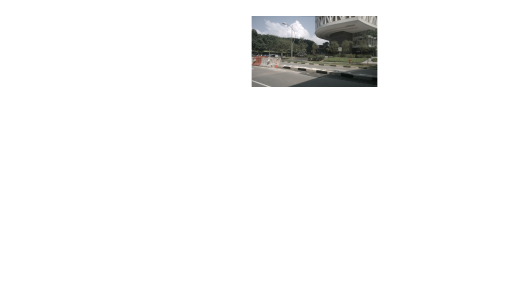 
    \caption{nuView Dataset sample generation with prompt.}
    \label{fig:nu-view-sample}
\end{figure}

\paragraph{nuCaption}
We train and evaluate our 3D captioning results on LiDAR-LLM's \cite{2023_Yang_LiDAR_LLLM_Exploring_the_Potential_of_Large_Language_Models_for_3D_LiDAR_Understanding} \textit{nuCaption} dataset\footnote{https://huggingface.co/datasets/Senqiao/LiDAR-LLM-Nu-Caption}. The dataset focuses on the traffic situation, which is crucial for prediction and planning in autonomous driving. For the training of BEV-LLM, we append positional information to the samples. 
\paragraph{nuView}
We propose the \textit{nuView} dataset containing textual descriptions corresponding to the six positional views from nuScenes, enabling to train context-aware models requiring multimodal positional encoding. Compared to nuCaption, the scene descriptions are generally more comprehensive about environmental context and contain scene attributes that LiDAR-LLM cannot capture with unimodal input, such as lighting and weather conditions. We generate the data using Llava-1.6 with a Llama-3-7B backbone to create captions for each of the six views of a nuScenes sample. An example is depicted in Figure~\ref{fig:nu-view-sample}, containing our \textit{view-specific} prompt appended to the instruction. 

\paragraph{GroundView} 
Further, we propose the \textit{GroundView} dataset, which originated from our internal research inspired by BEV-TSR~\cite{bevtsr2024}, containing precise captions based on the nuScenes sample annotations, from which natural language labels for grounding are created.
The dataset is a structured summary with predefined linguistic rules, covering environmental conditions and object occurrences from the combined views, distinguishing between \textit{one}, \textit{several} for two instances, and \textit{many} instances.

\subsection{Evaluation Metrics}

We evaluate the scene descriptions of BEV-LLM with standardized Natural Language Processing metrics. The BLEU~score~\cite{kishore2002} measures the precision of identical n-grams in the predicted text compared to the ground truth, with a BLEU score of~1 indicating identical texts. 
BLEU-$n$ describes n-gram precision comparing word sequences. BLEU-[1-4] scores measure the positional precision for one-, two-, three-, and four-word sequences.Rouge-L measures the longest common subsequence.
The BERT-score~\cite{Zhang2020} evaluates semantic similarity rather than literal word matching by comparing the contextual embeddings of the tokenized sequences.

\subsection{Qualitative Results}
\label{sec:qualitative}
\begin{figure*}[t]
    \centering
    \def\svgwidth{\linewidth}
    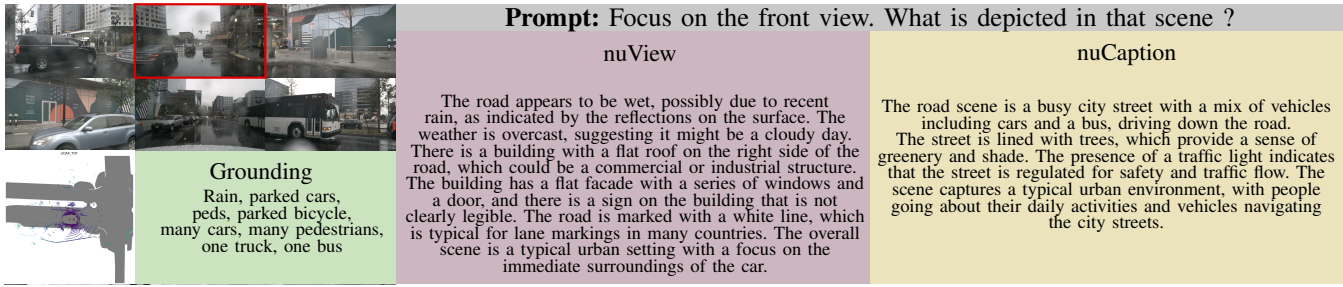 
    \caption{Prediction sample of BEV-LLM-1B using the various data source (GroundView, nuView and nuCaption) based on the shown prompting.}   
    \label{fig:qualitative_eval}
\end{figure*}

We present the qualitative results in Figure \ref{fig:qualitative_eval}. BEV-LLM-1B describes objects and movements in addition to the overall scene context, the environmental setting, and the infrastructure. In contrast to LiDAR-LLM, which exhibits illusions when describing aspects that are not captured with LiDAR as a single modality, such as weather conditions \cite{2023_Yang_LiDAR_LLLM_Exploring_the_Potential_of_Large_Language_Models_for_3D_LiDAR_Understanding}, BEV-LLM describes these comprehensively. Trained on nuView, BEV-LLM accurately describes weather, surrounding infrastructure, and environmental context, while the model trained on the nuCaption lacks these details. However, BEV-LLM is more accurate with vehicles in nuCaption, which offers more concise traffic-focused descriptions.
In both samples, BEV-LLM is prompted to focus on the front view, which is correctly captured by the model. Both models predict a correct set of occurrences in the scene and show their expected behavior. The nuView model correctly predicts the weather and road conditions, whereas the nuCaption sample detects a traffic light and focuses more on the traffic conditions. 

\begin{table}[t]
    \captionsetup{width=\columnwidth} 
    \LARGE
    \resizebox{\columnwidth}{!}{ 
    \begin{tabular}{|c|c|c|c|c|c|c|}
        \hline
        Models & B-1 ↑ & B-2 ↑ & B-3 ↑ & B-4 ↑ & BERT-P ↑ & Parameters \\ \hline
         Mini-GPT4 \cite{zhu2023minigpt} & 14.97 & 6.76 & 3.74 & 2.63 & 84.38 & 7B \\ 
        LLaVA1.5 \cite{liu2024improved} & 19.92 & 12.10 & 8.57 & 5.37 & 85.01 & 7.06B\\ 
        Instruct-BLIP \cite{2023_Li_BLIP_2_Bootstrapping_Language_Image_Pre_Training_with_Frozen_Image_Encoders_and_Large_Language_Models} & 18.67 & 13.38 & 7.41 & 5.20 & 85.89 & 7.91B \\ 
        LLama-AdapterV2 \cite{zhang2023llama} & 30.17 & 17.34 & 10.40 & 7.45 & 87.45 &7.22B \\
        BEV-LLM-8B (ours) & 39.51 & 29.34 & 23.90 & 19.87 & 86.28 &8.17B \\
        LiDAR-LLM  \cite{2023_Yang_LiDAR_LLLM_Exploring_the_Potential_of_Large_Language_Models_for_3D_LiDAR_Understanding} & \textbf{40.98} & 29.96 & 23.43 & 19.26 & \textbf{91.32} & 6.96B \\ 
        BEV-LLM-1B (ours) & 40.61 &\textbf{30.02}& \textbf{24.42} & \textbf{20.28} & 86.15 & \textbf{1.35B} \\
        \hline
    \end{tabular}
    }
    \caption{Performance metrics for 3D Captioning tasks across different models on the nuCaption dataset. BERT precision is used for comparability.}
    \label{tab:3d_captioning}
\end{table}

\subsection{Quantitative Results}
\label{sec:quantitative}
\subsubsection{3D-Scene Captioning on nuCaption}
We evaluate BEV-LLM on the nuCaption dataset and compare it to other state-of-the-art methods for 3D captioning (see Table~\ref{tab:3d_captioning}). For comparability, we use the BERT precision metric.
The number of input queries (512) is similar to LiDAR-LLM (576), both with 768 query dimensions.
BEV-LLM-1B achieves best performance in the BLEU-2, 3, and 4 scores exceeding the second-best model, LiDAR-LLM, by 5\% with a BLEU-4 score of 20.28\% and running second on BLEU-1 with 40.98\% being 1\% behind LiDAR-LLM. These results suggest that our model contains more coherent multi-word sequences matching the ground truth than the baselines. 
On the BERT score, BEV-LLM yields results below LiDAR-LLM and LLama-AdapterV2. This result paired with the high BLEU score precision indicates that the BERT score is influenced by the rich generation sequences of BEV-LLM.
Furthermore, BEV-LLM-1B supersedes BEV-LLM-8B in all BLEU scores but exhibits slightly worse BERT performance. We explain the better word matching by the higher number of input queries used for BEV-LLM-1B.

\subsubsection{3D-Scene Captioning on nuView}\begin{table}[h]
    \LARGE    
    \resizebox{\columnwidth}{!}{
    \begin{tabular}{|c|c|c|c|c|c|c|c|c|}
        \hline
        Dataset & Models & B-1 ↑ & B-2 ↑ & B-3 ↑ & B-4 ↑ & BERT-F1 ↑ & Rouge-L-F1 ↑ & Params \\ \hline
        \multirow{2}{*}{NuView } & BEV-LLM-1B & \textbf{49.40} & \textbf{35.01} & \textbf{26.74}  & 20.96 &\textbf{85.67} & \textbf{38.25}  & 1.35B \\
        & BEV-LLM-8B & 49.08 & 34.80 & 26.71 & \textbf{21.04} & 85.55& 37.62&8.17B\\
        \hline
        \multirow{2}{*}{GroundView} & BEV-LLM-1B & \textbf{28.73} & \textbf{17.83} & \textbf{11.24} & \textbf{6.87} & \textbf{89.04} & 36.74 & 1.35B \\
        & BEV-LLM-8B & 26.87 & 17.53  & 11.02 & 6.83& 88.60 & \textbf{37.59} & 8.17B \\
        \hline
    \end{tabular}
    }

    \captionsetup{width=\columnwidth}
    \caption{Performance metrics for 3D Captioning tasks across our two model configurations using our nuView and object-focused GroundView dataset. Measured with BERT F1 combining precision and recall.}
    
    \label{tab:3d_captioning_nuView}
\end{table}

Table \ref{tab:3d_captioning_nuView} presents ablation studies comparing BEV-LLM-1B and BEV-LLM-8B on our nuView dataset. BEV-LLM-1B slightly outperforms the 8B model in BLEU-1,2 and 3 scores with decreasing distance. BEV-LLM-8B yields better results on BLEU-4 with 21.04\%, thus containing more multi-word n-grams compared to text length. Coupled with the higher BERT and Rouge-L scores, the results suggest that the predictions of BEV-LLM-1B are generally more concise and semantically on point. The nuView dataset will be released with initial results as a benchmark for further research.

\subsubsection{Object grounding on GroundView}
The ablation studies on the GroundView dataset are depicted in the bottom row of Table \ref{tab:3d_captioning_nuView}. BEV-LLM-1B shows improved performance in all metrics but the Rouge-L score compared to the 8B parameter model, suggesting that the higher query number overweighs the larger model size. The GroundView descriptions are generally shorter than the descriptions from nuView, indicating more concise predictions from the 1B model. Compared to the results on nuView, BEV-LLM has a worse BLEU score on these compact descriptions, which are largely comma separated numerations of objects, with only 6.87\% on BLEU-4, but has a higher BERT score of 89.04\%, suggesting that the model outputs more expressive language descriptions while capturing the semantic meaning of the ground truth well.
The GroundView dataset will be released with initial results as a benchmark for further research.

\section{DISCUSSION}
\label{sec:discussion}
Our experiments demonstrate that BEV-LLM delivers comprehensive scene descriptions with contextual information and competes with state-of-the-art models, despite having only 1.35B parameters. By processing multimodal sensor inputs over BEV, it integrates data specific to individual sensors, such as atmospheric information (e.g., weather data) from cameras, alongside detailed infrastructure and object data. This capability is particularly valuable for training reasoning LLMs to draw context-based conclusions, thereby enhancing both safety and trust.
Like all LLMs, BEV-LLM is prone to hallucinations. While our model achieves the best results on BLEU scores for word sequences, it shows a performance drop for the BERT~score. This could be due to the model’s focus on generating fluent, contextually appropriate text that may differ slightly in phrasing from the reference captions, affecting the BERT~score, which relies on precise semantic alignment. 

Our proposed dataset, nuView, currently lacks accurate grounding information. By extending the GroundView dataset, we target a better object-centric reasoning and 3D grounding, while maintaining rich contextual scene information from nuView and nuCaption, aiming to improve both the semantic accuracy and thus the model’s performance on the BERT~score. Furthermore, human annotation can increase the diversity and completeness of the dataset descriptions.
We further plan to incorporate temporal dynamics, enabling models to generate descriptions that reflect changes over time and thus allow the tracking of object movements and environmental shifts.

\section{CONCLUSION}
\label{sec:conclusion}
Scene descriptions are crucial for interpretable decision-making and reasoning in LLMs, ensuring safety and transparency in autonomous driving. With advancements in BEV maps from LiDAR and camera fusion, accurate object positioning and detailed attribute descriptions are now possible.
We introduced BEV-LLM, a lightweight model for 3D captioning of autonomous driving scenes. BEV-LLM leverages BEVFusion to combine 3D LiDAR point clouds and multi-view images, incorporating a novel absolute positional encoding for view-specific scene descriptions. Despite using a small 1B parameter base model, BEV-LLM reaches a competitive performance, surpassing state-of-the-art methods in BLEU-2, 3, and 4 scores. Additionally, we released two new datasets — nuView and GroundView — along with initial benchmarking results.

\section*{ACKNOWLEDGMENT}
We thank Viktor Kilic for contributing the GroundView dataset.

This work is a result of the joint research project STADT:up (grant no. 19A22006N). The project is supported by the German Federal Ministry for Economic Affairs and Climate Action (BMWK), based on a decision of the German Bundestag. The authors are solely responsible for the content of this publication. 
\bibliographystyle{plain}
\bibliography{references.bib}

\end{document}